\documentclass[a4paper]{article}
\usepackage{iwslt18-submission,epsfig}
\usepackage{cite}
\usepackage{amsmath,amssymb}
\usepackage{algorithmic}
\usepackage{graphicx}
\usepackage{times}
\usepackage{url}
\usepackage{bm}
\usepackage{multirow}
\usepackage{caption}
\usepackage{subcaption}
\usepackage{dblfloatfix}
\usepackage{listings}
\usepackage[utf8]{inputenc}
\usepackage[vietnamese,english]{babel}
\usepackage{CJKutf8}

\usepackage[T1]{fontenc}
\usepackage{color}

\usepackage{authblk}

\def\reg{{\rm\ooalign{\hfil
     \raise.07ex\hbox{\scriptsize R}\hfil\crcr\mathhexbox20D}}}

\newcommand{\JP}[1]{\begin{CJK}{UTF8}{min}#1\end{CJK}}

\newcommand*{\thead}[1]{\multicolumn{1}{|c|}{\bfseries #1}}

\def\BibTeX{{\rm B\kern-.05em{\sc i\kern-.025em b}\kern-.08em
    T\kern-.1667em\lower.7ex\hbox{E}\kern-.125emX}}
    
\newcommand{\vect}[1]{\mathbf{#1}}

\title{How Transformer Revitalizes Character-based Neural Machine Translation:\\An Investigation on Japanese-Vietnamese Translation Systems}

 \makeatletter
 \def\name#1{\gdef\@name{#1}}
 \makeatother
 \name{{\em Thi-Vinh Ngo$^1$, Thanh-Le Ha$^2$, Phuong-Thai Nguyen$^3$, Le-Minh Nguyen$^4$}}

 \address{$^1$University of Information and Communication Technology, TNU, Vietnam \\
 $^2$Institute of Anthropomatics and Robotics, KIT, Germany \\
 $^3$University of Engineering and Technology, VNU, Vietnam \\
 $^4$School of Information Science, JAIST, Japan \\
 {\small \tt ntvinh@ictu.edu.vn,thanh-le.ha@kit.edu,thainp@vnu.edu.vn,nguyenml@jaist.ac.jp}
 }

\begin{document}
\maketitle
\begin{abstract}
While translating between East Asian languages, many works have discovered clear advantages of using  characters as the translation unit. Unfortunately, traditional recurrent neural machine translation systems hinder the practical usage of those character-based systems due to their architectural limitations. They are unfavorable in handling extremely long sequences as well as highly restricted in parallelizing the computations. In this paper, we  demonstrate that the new transformer architecture can perform character-based translation better than the recurrent one. We conduct experiments on a low-resource language pair: Japanese-Vietnamese. Our models considerably outperform the state-of-the-art systems which employ word-based recurrent architectures.

\end{abstract}

\section{Introduction}


Neural machine translation (NMT) has achieved the state-of-the-art performances in recent machine translation campaigns for many language pairs due to its fluent and accurate outputs\cite{wu2016google,cettolo2016iwslt,cho2016adaptation}. Yet often those outputs from NMT are \textit{over-translated}, where words or phrases are redundantly repeated in the translations, thus affecting their readability. One reason leading to \textit{over-translation} in NMT is, unlike the traditional statistical machine translation (SMT), basic NMT architectures and their beam search do not explicitly model coverage. Furthermore, their content-based attention mechanism, which effectively aligns an arbitrary number of source words to the only target word at a time when decoding without any length control, intensifies the problem. Consequently, the best and simplest strategy to avoid \textit{over-translation} when translating between two languages having lots of length mismatches might be to segment the texts in some way so that the systems could work with \textit{one-to-one alignments} as many as possible\footnote{By terming \textit{one-to-one alignment}, we mean that one translation unit of the source sentence  corresponds to one translation unit of the target sentence and vice versa.}. On the other hand, given sufficiently large data, a well-designed NMT architecture is capable of automatically learning good alignments with its attention mechanism.

For translation systems between Romance, Balto-Slavic and Germanic languages, unsupervised subword segmentation methods, such as Byte-Pair Encoding\cite{Sennrich2016a}, are often used and they show great improvements. Subword-based translation systems possess two important advantages. First, they reduce the vocabulary size, hence, making the model memory and performance-efficient. Second, they are able to competently deal with unknown and rare words. For many other languages in East, South and South-East Asia, however, the preprocessing frequently requires more complicated and expensive supervised tokenization methods to segment the texts into decent translation units, since in those languages, word boundaries are not signaled by the whitespaces. 

In order to liberate the whole translation system from the dependencies on those non-trivial tokenization methods (including subword segmentation) while still effectively dealing with the out-of-vocabulary (OOV) problem, character-level approaches\footnote{Here we mean the pure character-level, in which the basic translation unit is a character, not the subword or the n-gram-character ones.} have been investigated. However, those approaches have not enjoyed much success in machine translation as when they are applied in other natural language processing tasks, since it is difficult  for the conventional recurrent-based neural translation architecture to properly model the relationship between characters and their meanings due to the inherent limitations of recurrent neural networks (RNN) in handling long sequence.

A more-recently-proposed neural machine translation architecture, the transformer, owns an important characteristic: via its self-attention blocks, it allows modeling arbitrarily-long-distance relationships with a constant number of operations during training. Thus, the transformer could alleviate the limitations of RNN-based architectures and make the character-level translation become effective and practical. 

The paper is structured as follows. We start with a detailed discussion about character-level translation in comparison with word and subword counterparts (Section~\ref{chatran}). Then we revise the recurrent architecture as well as the transformer architecture in Section~\ref{architecture}. Section~\ref{experiments} describes our experiments and our analysis about how the transformer revitalizes character-based machine translation.  Finally, the paper ends with conclusion and future work.

\section{Character-Level Translation}
\label{chatran}
In general, most of the neural machine translation systems prefer word and subword as the translation unit than character due to the following limitations of character-level translation approach:
 
\begin{enumerate}
\item Unlike words or subwords that bear some meaning, a character in many languages simply represents an orthographical symbol, and the relationship between that symbol and its meaning are arbitrary from the linguistic point of view. For example, it is unfeasible to induce any meaning from the English character "c" or the Japanese \textit{hiragana} character \JP{ガ} when they stay alone.

\item Since the sequence lengths when the translation unit is character are 3 to 10 times bigger than those of word-based translation\footnote{This factor depends on the languages.}, the neural architecture needs to be able to capture longer-distance dependencies in order to produce reasonable translation. However, the recurrent architectures often fail to handle such long-term dependencies even when they employ gated recurrent units like long short-term memories (LSTMs) or gated recurrent units (GRUs).

\item The same reason prohibits the usage of recurrent architectures in practice. The number of model parameters increases proportionally with the number of time steps (i.e. the sequence length). Thus, the training and inference processes are slowed down as well as the demand of memory footprint escalates just to get a similar modeling power on the same sentence being represented as a sequence of words or subwords.
\end{enumerate}

In contrast, there are handful of languages in which character-based translation is clearly favorable. Chinese, for example, is a logo-syllabic language where the graphemes represent morphemes. Each concept in Chinese often comprises of two characters and each of those characters bears some meaning (\textit{morpheme} in morphology), similar to the amount of semantic information that a subword or even a word conveys in other languages. For instance, the word ``\JP{遊歷}'' (travel) is constituted by two characters ``\JP{遊}'' (move) and ``\JP{歴}'' (experience), meaning ``moving and experiencing''. This holds true with Japanese \textit{kanji} scripts\footnote{\textit{Kanji} means \textit{Chinese character} and a large number of \textit{kanji} characters are the same as their Chinese counterpart.}, one among three scripts in Japanese (\textit{kanji}, \textit{hiragana} and \textit{katakana}). A Vietnamese word, on the other hand, is written as a sequence of Latin characters and white-spaces, thus, each character hardly carries any meaning as in case of other Latin-based languages. However, if we consider each morpheme, which is separated from others by white-spaces, to be a Chinese-like character as there are almost one-to-one mappings between a morpheme in Vietnamese and a character in Chinese, character-based approaches work almost analogously between the two languages. Let us take the Vietnamese correspondence of the Chinese word ``\JP{遊歴}'' above: ``\foreignlanguage{Vietnamese}{du lịch}'' (travel) is made up of from two morphemes ``\foreignlanguage{Vietnamese}{du}'' (move) and ``\foreignlanguage{Vietnamese}{lịch}'' (experience), while each morpheme is a sequence of Latin-based characters. \textit{Due to this reason, in the context of this work,  from here onwards we would like to consider a \textbf{Vietnamese morpheme} a \textbf{character}}. Table~\ref{tab:example1} shows some examples of such character-based mappings among Chinese, Japanese  \textit{kanji} and Vietnamese. The mappings help when translating between those languages at the character level. 
 
\begin{table*}[!h]
  \begin{center}
    \begin{tabular}{|l|l|l|} 
    \hline \hline
      \thead{Japanese kanji \& Chinese} & \thead{Tokenized Japanese kanji} & \thead{Vietnamese}\\
      \hline
      ``\JP{校長}(headmaster)'' & ``\JP{校}(school)'' and ``\JP{長}(head/lead)'' & ``\foreignlanguage{Vietnamese} {hiệu trưởng}'' (``\foreignlanguage{Vietnamese}{hiệu}''$\equiv$``\JP{校}'', ``\foreignlanguage{Vietnamese}{trưởng}''$\equiv$``\JP{長}'')\\
      \hline
      ``\JP{村民}(villager)'' & ``\JP{村}''(village) and ``\JP{民}''(citizen) & ``\foreignlanguage{Vietnamese} {dân làng}'' (``\foreignlanguage{Vietnamese}{dân}''$\equiv$``\JP{民}'', ``\foreignlanguage{Vietnamese}{làng}''$\equiv$``\JP{村}'')\\
      \hline
      ``\JP{同時}(simultaneous)'' & ``\JP{同}(same)'' and ``\JP{時}(time)'' & ``\foreignlanguage{Vietnamese} {đồng thời}'' (``\foreignlanguage{Vietnamese}{đồng}''$\equiv$``\JP{同}'', ``\foreignlanguage{Vietnamese}{thời}''$\equiv$``\JP{時}'')\\
      \hline
      ``\JP{日本人}'' (Japanese people) &  ``\JP{日本}''(Japan) and ``\JP{人}''(human)  & ``\foreignlanguage{Vietnamese} {người Nhật}'' (``\foreignlanguage{Vietnamese}{người}''$\equiv$``\JP{人}'', ``\foreignlanguage{Vietnamese}{Nhật}''$\equiv$``\JP{日}'')\footnotemark\\
      \hline \hline 
    \end{tabular}
  \end{center}
  \vspace*{-0.3cm}
  \caption{\label{tab:example1}Some examples of words in Japanese \textit{kanji} (Chinese characters) and their character mapping in Vietnamese.}
\end{table*}

\footnotetext{``\foreignlanguage{Vietnamese} {người Nhật}'' (Japanese people) is the short form but a more popular version of ``\foreignlanguage{Vietnamese} {người Nhật Bản}'' (``\foreignlanguage{Vietnamese}{người}''$\equiv$``\JP{人}'', ``\foreignlanguage{Vietnamese}{Nhật}''$\equiv$``\JP{日}'', ``\foreignlanguage{Vietnamese}{Bản}''$\equiv$``\JP{本}'').}

A unique characteristic of Japanese compared to Chinese and Vietnamese is that in written Japanese, \textit{kanji}, \textit{hiragana} and \textit{katakana} scripts are often mixed. As a consequence, when translating from Vietnamese or Chinese to Japanese, each source character is mapped to a \textit{kanji} character or to a short sequence of \textit{hiragana} or \textit{katakana} characters in the target side. To avoid this length mismatch problem which affects greatly to recurrent architectures, the studies opt to either word-based translation using tokenization methods or sub-character translation. To the best of our knowledge, state-of-the-art translation systems from Chinese and Vietnamese to Japanese follow those two directions in a recurrent framework instead of utilizing pure charecter-based approaches. The first direction, word-based approaches, requires good supervised tokenization or word segmentation tools which might be expensive for some languages and domains\cite{zhangkomachi2018}. The later demands external knowledge resources of sub-character level, such as Chinese character radicals or how to convert from a character to several strokes\cite{Zhang2018}. Furthermore, this direction is not applicable for non-logographic languages like Vietnamese or Korean. 

In this work, we seek for a simple solution featuring character-level translation, since it does not require external tools and resources, but capable of alleviating the typical difficulties of long distance modeling, both in theory and practice, of recurrent-based architectures. We hypothesized that transformer architecture is well suited to address or at least reduce the problem, thus, improves the bar performance of character-level translations between Japanese and other languages.

\section{Neural Machine Translation Architectures}
\label{architecture}
In this section, we describe two NMT architectures which are the most popular instances of the general neural encoder-decoder framework applied in sequence-to-sequence problems: recurrent-based and transformer.

Given a source sentence $\vect{x} = \{x_1,..., x_n\}$, the encoder is a neural architecture which reads every words $x_i$ and encodes a representation of the sentence into a fixed-length vector, called the \textit{context vector}. The context vector is often time-specific, representing the source sentence at different time steps. It is calculated via \textit{attention mechanism}, which is essentially a weighted combination of the source hidden states $\vect{h}_i$. The decoder, which is another neural architecture, generates one target word every time step to form a translated target sentence $\vect{y} = (y_1,..., y_m)$ in the end.  In addition to the information from previous generated sequence $\vect{y}_{1:j-1} = (y_1,..., y_{j-1})$, the decoder is also conditioned on the context vector $\vect{c}_j$, which contains the source sentence information, to produce the next target word $y_j$ at the time step $j$. In practice, this is modeled a probabilistic distribution over the target vocabulary by applying a softmax layer on the decoder representation $\vect{z}_{j}$:
$$ p(y_j|\vect{y}_{1:j-1}, \vect{c}_j) = \text{Softmax}(W\vect{z}_{j}+b) $$


The main difference between the recurrent-based and the transformer is  how the encoder and decoder model the sequence, which is shortly described in the section below.

\begin{figure*}[t]
\begin{center}
\includegraphics[scale=0.45]{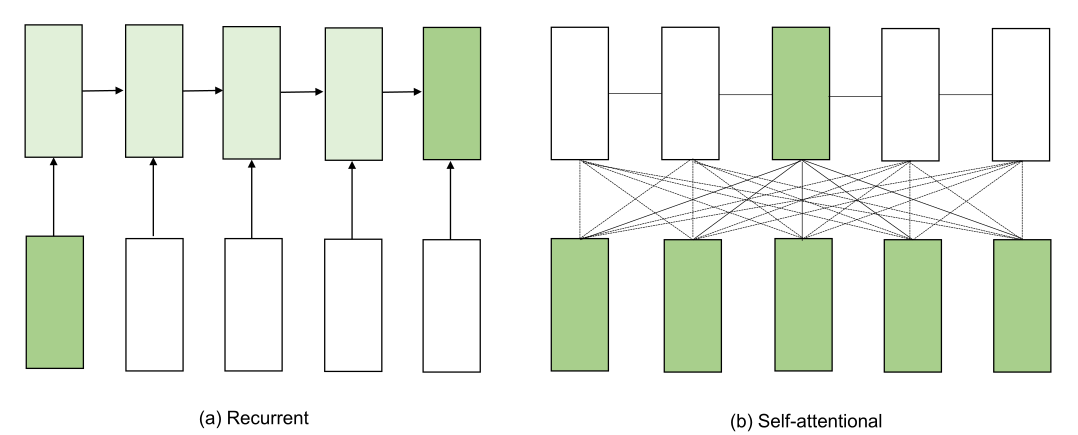}
\caption{\label{fig:RNNTrans} Recurrent and self-attention architectural differences.}
\end{center}
\end{figure*} 

\subsection{Recurrent Architecture}
As the name suggested, recurrent architectures employ recurrent-based units as the main part of its encoder and decoder. In the encoder, the hidden state $\vect{h}_i$ is modeled by a bidirectional recurrent unit (e.g. LSTM\cite{HochreiterLSTM} or GRU\cite{Cho2014}), taking into account the current word's embeddings $\vect{s}_i$ and the hidden state of the previous word $\vect{h}_{i-1}$. $\vect{h}_i$ encodes the source sentence up to the time $i$ from both forward and backward directions:
\[
\begin{aligned}
& \vect{h}_i=[\overrightarrow{\vect{h}}_i,\overleftarrow{\vect{h}}_i] \\
& \overrightarrow{\vect{h}}_i=\text{Recurrent}(\overrightarrow{\vect{h}}_{i-1},\vect{s}_i) \\
& \overleftarrow{\vect{h}}_i=\text{Recurrent}(\overleftarrow{\vect{h}}_{i+1},\vect{s}_i) \\
\end{aligned}
\]

Similarly the decoder uses recurrent units to calculate the target hidden state $\vect{z}_j$ based on the previous hidden state of the decoder $\vect{z}_{j-1}$, the embeddings of the previous target word $\vect{t}_{j-1}$ 
and the time-specific context vector $\vect{c}_j$:
\[
\begin{aligned}
& \vect{z}_{j}=\text{Recurrent}(\vect{z}_{j-1}, \vect{t}_{j-1}, \vect{c}_j) \\
\end{aligned}
\]
In recent NMT architectures, the encoder and decoder are constructed by stacking several recurrent layers, and residual connections\cite{he2016deep} are added between layers in order to make the training of the deep network feasible. The attention mechanism is originally applied in the recurrent-based architecture, as mentioned before, then plays a more important role in the transformer architectures.

\subsection{Transformer Architecture}
Transformer architectures based on the concept of \textit{attention}, which is a generalized version of the attention mechanism used in recurrent architectures\footnote{More precisely, the attention mechanism mentioned here is the generalized version of the \textit{dot} attention\cite{Luong2015b}, which is the most popular implementing way of attention.}:

\begin{equation}
\begin{aligned}
\text{Attention}(\vect{Q},\vect{K},\vect{V}) =  \text{Softmax}(\displaystyle \frac{\vect{Q}\vect{K}^T}{d}\vect{V})  \\
\end{aligned}
\label{eq:att}
\end{equation}
where $d$ is the scaling factor, depending on the size of the input to that attention layer.

Basically, this attention mechanism models the relationships between \textit{queries} $\vect{Q}$ and tuples of \textit{keys} and \textit{values} ($\vect{K}$,$\vect{V}$). In the original attention used in between the encoder and the decoder of recurrent architectures (and also between those of the transformer architectures), the queries come from the decoder's hidden states, and the keys, that each of them is also the corresponding value, are all the hidden states coming from the encoder. The transformer, however, also features a special kind of attention in its encoder and decoder, called \textit{self-attention}. In self-attention encoders, the queries, keys and values all come from the representation of the source sentence. This allows each position attend to every other position, automatically figuring out some relationship among source words. Similarly, \textit{self-attention} is applied in the decoder, with a small modification: the future positions are masked out since the future information (future target words) are not available at the inference time. 

Transformer architectures employ \textit{multi-head attention}, each head is the result of Formula~\ref{eq:att}, and each head models a relationship among source or target sentences. They are then concatenated and linearly combined into the multi-head attention. Due to the fact that all of the attention heads and multi-head attentions are calculated by feed-forward layers, parallel calculations of the whole architecture\footnote{Transformer consists of several stacked encoder and decoder blocks. In each encoder or decoder block, besides self-attention layers, there are position-wide feed-forward layers as well as residual and normalization layers. Since the encoder and decoder using self-attention does not explicitly encode the information of the sequence order like the recurrent ones, a positional encoding is injected along the word embeddings of both the encoder and decoder. For more details, please refer to~\cite{VaswaniSPUJGKP17}.} across time steps is straightforward, constant to the length of the sequence. Furthermore, each state in the self-attention encoder or decoder is connected directly to all other states, no matter how far in the order they are introduced. In other words, long distance relationships are modeled better in self-attention mechanism than in recurrent architectures which rely on forget mechanism. So multi-head self-attention in theory allows us to model various aspects of the extremely long source and target sequences. In practice, the context that self-attention can effectively model is often beyond every sentence. This is the key answer for what we questioned our character-based translation systems in Section~\ref{chatran}. Figure\ref{fig:RNNTrans} summarizes the main architectural differences between the recurrent-based and the transformer.

\subsection{Transformer vs. Recurrent in Character-based MT}
With the differences between two architectures, we hypothesize that transformer could address the problems that character-based recurrent translation systems encounter. Specifically, transformer is expected to offer more benefit than the recurrent in those aspects:

\textbf{Jointly Learn Tokenization and Representation.} The most complicated recurrent unit, LSTM, has three different gates: input, output and forget gates, thus, it possesses excellent memory mechanism. It is still unable, however, to jointly learn word segmentation or tokenization and the relationship between two words. On the other hand, the transformer can have one attention head to learn how to combine possible characters, even they are not consecutive, into a meaning unit and other heads to learn different dependencies among words in the sequence. Another possible scheme is that each head in the multi-head attention can learn how to combine characters in a specific way suitable for learning a specific relationship in the sequence.   

\textbf{Long-distance Modeling.} Transformer is better to capture information of long-distance dependencies in a sequence than the recurrent one. While the relationship of two words far from each others is modeled correspondingly far in the recurrent, which is extremely difficult to be learned, that relationship is directly modeled in the self-attention regardless of the distance between them. A sentence in Japanese would be three or four times longer if it is represented as a sequence of characters compared to that as a sequence of words, and that factor is around two times in Chinese and Vietnamese.  

\textbf{Highly Parallelization.} Transformer allows parallel computations not only over the stacked layers but also across the time steps. In training, the number of computing operations in each layer of transformer is constant whereas they are proportional to the sequence length in case of the recurrent architecture. With the same training time, a transformer can have much larger modeling capacity than a recurrent-based model. Again, it is more effective and efficient in translating a longer sequence of characters than a sequence of words.

\begin{center}
\begin{table*}[ht]
\vspace*{-0.1cm}
{\small
\hfill{}
\begin{tabular}{|l|l|c|c|c|c|c|c|c|c|}
\hline \hline
\multicolumn{6}{|c|}{ \textbf{Japanese$\Rightarrow$Vietnamese}} \\ 
\hline
\multirow{1}{*}{} & \multirow{1}{*}{System}& BLEU & $\Delta$ BLEU & RIBES & $\Delta$ RIBES \\ \hline
 (1) & Word2WordRecurrent & 11.05 & -2.29 & 0.663 & -0.025 \\
 (2) & Word2WordTransformer & 11.72 & -1.62 & 0.681 & -0.007 \\ \hline
 (3) & Char2CharRecurrent & 10.06 & -3.28  & 0.657 & -0.031\\
 (4) & Char2CharTransformer & \textbf{13.34} & -  & \textbf{0.688} & -\\
\hline
\hline

\multicolumn{6}{|c|}{ \textbf{Vietnamese$\Rightarrow$Japanese}} \\ 
\hline
\multirow{1}{*}{} & \multirow{1}{*}{System}& BLEU & $\Delta$ BLEU & RIBES & $\Delta$ RIBES \\ \hline
 (1) & Word2WordRecurrent & 11.13 & -3.92 & 0.593 & -0.098 \\
 (2) & Word2WordTransformer & 13.07 & -1.98 & 0.679 & -0.012 \\ \hline
 (3) & Char2CharRecurrent & 9.61 & -5.44  & 0.566 & -0.125\\
 (4) & Char2CharTransformer & \textbf{15.05} & -  & \textbf{0.691} & - \\
\hline
\\
\hline
\end{tabular}}
\hfill{}
\caption{\label{tab:ViJa} {Comparisons of Japanese$\Leftrightarrow$Vietnamese translation systems' results on {\tt tst2010}.}}

\end{table*}
\end{center}
\vspace*{-1.5cm}
\section{Experiments and results}
\label{experiments}

We would like to verify our hypotheses with Japanese$\Leftrightarrow$Vietnamese translations. More specifically, we set up the following experiments and conduct the comparisons among them:

\textbf{Word2WordRecurrent}: Word-based translation system using \textit{recurrent architecture} with the \textit{best tokenizations}. Following the description of \cite{VinhNgo2018}, the best system performs tokenization and sub-word segmentation for Japanese and the modified sub-word segmentation for Vietnamese. However, in their paper, they also mentioned that the system using supervised word segmentation achieved similar result to their modified sub-word segmentation for Vietnamese and we decided to take the supervised word segmentation system since it is easier to replicate that system.   

\textbf{Word2WordTransformer}: Word-based translation system using \textit{transformer architecture} with the \textit{best tokenizations} (the same methods applied in the first system).

\textbf{Char2CharRecurrent}: Character-based system using \textit{recurrent architecture} \textit{without any tokenization}. Note that on the Vietnamese side, \textbf{\textit{character}} here means \textbf{\textit{morpheme}}, separated to others by white-spaces. Please refer to Section~\ref{chatran}. 

\textbf{Char2CharTransformer}: Character-based system using \textit{transformer architecture} \textit{without any tokenization}.

\subsection {Data} 
We use four Japanese-Vietnamese parallel corpora collected from various sources: (1) TED talks corpus in \cite{VinhNgo2018}, (2) Asian Language Treebank corpus in \cite{Riza2016}, (3) we extracted bilingual sentences from the multilingual Tatoeba corpus\footnote{\url{https://tatoeba.org}}, and (4) we crawled examples of bilingual sentences from Glosbe\footnote{\url{https://en.glosbe.com/}}, an open multilingual online dictionary. After removing duplicate lines and filtering noisy data in (4) we obtained 210K sentence pairs\footnote{The compilation of the corpora is available at \url{https://github.com/ngovinhtn/JaViCorpus}.}. 
To evaluate the translation quality on several translation systems and also to compare with the first published Japanese$\Leftrightarrow$Vietnamese translation systems~\cite{VinhNgo2018}, we use {\tt dev2010} as the validation set  and  {\tt tst2010} for testing in all experiments. {\tt dev2010} and  {\tt tst2010} are sentences extracted from TED talks, thus, we can consider (1) is the in-domain training data. Comparing to \cite{VinhNgo2018}, validation and test sets are cleaned up to make sure that the length of all sentences do not exceed 100 tokens. 


  \subsection{Preprocessing} 
   We used {\tt kytea}\footnote{\url{http://www.phontron.com/kytea/}}~\cite{Neubig2011} to tokenize Japanese texts and then segmented into sub-words using BPE method~\cite{Sennrich2016a}. For Vietnamese texts, we first normalized them using the Moses scripts and then we employed {\tt pyvi}\footnote{\url{https://pypi.org/project/pyvi/}} to conduct supervised word segmentation on those texts.

  \subsection{System Architectures}
  
  ~~~~~~\textbf{Recurrent systems.} 
 We implement all recurrent translation systems using {\tt OpenNMT-py}  \footnote{\url{https://github.com/OpenNMT/OpenNMT-py}}\cite{opennmt}. In our models, the encoder is a bi-directional LSTM which has two layers and the decoder is another recurrent architecture with two LSTM layers, the hidden size for each layer is 512 dimension. The embedding size on both source and target is also 512. We use Adam optimizer to update weights with the learning rate is initialized at $0.001$ and then annealing on training. The size of each mini batch is 32 and the number of training epochs for each system are 15. Other parameters are the defaults of {\tt OpenNMT-py}. For each system, we choose the best model in terms of the accuracy on validation set.
 
  \textbf{Transformer systems.} 
  We employ the framework {\tt NMTGMinor}\footnote{\url{https://github.com/quanpn90/NMTGMinor}}, a variant of Transformer described in \cite{VaswaniSPUJGKP17}. For all our model, we use a stack of 4 layers for both encoder and decoder with the sizes of hidden units and embedding for each layer are the same as 512 and the number of heads are $H=8$. The size of inner feed forward layer is 1024.  The number of words on each mini batch are 4096 tokens. We use the scaling factor for all dropout layers is $0.2$ except on embedding indices is $0.1$. Like recurrent models, we also use Adam optimizer to learn weights but we initialize the learning rate at $1.0$ and do not use annealing on training. The output of loss function is smoothed with the factor of $0.1$. We train all of translation systems for 50 epochs and obtain the best model which have the smallest perplexity on the validation set.
  
\subsection{Results} 
We evaluate the quality of our translation systems using two measures including {\tt multi-BLEU} and {\tt RIBES}\footnote{\url{http://www.kecl.ntt.co.jp/icl/lirg/ribes/}} \cite{Isozaki2010}. Table~\ref{tab:ViJa} shows the results for Japanese$\Rightarrow$Vietnamese and Vietnamese$\Rightarrow$Japanese. To have more exact evaluation in case of Vietnamese$\Rightarrow$Japanese direction, we re-apply {\tt kytea} tokenization on the translated outputs, and calculate {\tt multi-BLEU} and {\tt RIBES} scores on these tokenized texts with the human references\footnote{All the recipes for those experiments are available at \url{https://github.com/ngovinhtn/charTransform}.}.

\textbf{Recurrent systems.} As we already analyzed in Section~\ref{chatran} and Section~\ref{architecture}, character-based translation systems employing recurrent architecture would suffer from a longer sequence given the same amount of information. Furthermore, it is difficult for the recurrent architecture to jointly learn tokenization and translation. Using the same recurrent architecture, the character-based system performs 1.44 and 1.09 BLEU scores less than the word-based system using the best tokenization methods in Japanese$\Rightarrow$Vietnamese and Vietnamese$\Rightarrow$Japanese, respectively.  

\textbf{Word2Word systems.} We observe the fact that using transformer architectures in word-based systems brings improvements in both of the translation directions, especially in Vietnamese$\Rightarrow$Japanese (BLEU improvement is 1.94, RIBES improvement is 0.086). It might reflect the better modeling capacity of the transformer over the recurrent.  

\textbf{Character-based transformer systems.} Those systems achieve best results on both directions, significantly outperform the best systems reported in \cite{VinhNgo2018}, which are the word-based systems using the best tokenization methods (BLEU scores 13.34 vs. 11.05 on Japanese$\rightarrow$Vietnamese and 15.05 vs. 11.15 on Vietnamese$\rightarrow$Japanese). Moreover, character-based systems do not require external knowledge or tools in order to perform tokenization. This advantage makes the translation systems more scalable and applicable in new domains and in similar languages where such tools and knowledge do not exist or expensive and difficult to create. Our systems set a new state-of-the-art results on Japanese$\Rightarrow$Vietnamese and Vietnamese$\Rightarrow$Japanese translations. The results confirms our hypotheses about the superiors of using transformer architectures on character-level translation.

\section{Related Works} 
To alleviate the weakness of word-based translation models, many works recently have inspected translation tasks on several levels like sub-word units or characters. \cite{Sennrich2016a} proposed BPE algorithm for learning rules to convert a word into sub sequences. \cite{VinhNgo2018} developed a segmentation method for Vietnamese. These approaches are unsupervised ways. \cite{Lee2016} have investigated translation systems at entirely character level on many Indo-European language pairs on sequence-to-sequence models using recurrent architecture. To dealing with the unfavorable impacts of long-term dependencies on translation systems, they add some more layers to the encoder to obtain a shorter representation from the input sentence. However, their experiments have shown these efforts still fail to solve this problem. \cite{Cherry2018} have also revisited character-based translation systems using RNNs but they extended their systems with a component for compressing of character sequences to speed up computation and a hierarchical multi-scale LSTM network for handling length sentences. \cite{Zhang2018} have achieved improvements on Japanese$\Rightarrow$Chinese translation task based on the survey of character translation. As previous woks, they also employ traditional NMT system with recurrent encoder-decoder and attention mechanism. In addition, they convert all katakana characters, Arabic numerals, Latin symbols and other special symbols in Japanese texts to new symbols corresponding. For these reason, many symbols in Japanese are  derived from Chinese symbols. This changes the original Japanese texts and makes the systems are more difficult to do translation in the reverse direction as Chinese$\Rightarrow$Japanese. Briefly, all of above works has conducted their experiments on recurrent architecture and modified their architectures or perform some replacing operators in the preprocessing.
\section{Conclusion}
We have investigated character-based NMT systems using transformer and compared them to the state-of-the-art word-based recurrent systems. Our results have shown that transformer is capable of learning long-term dependencies, so they can translate better at character level. In the future, we would like to exploit our systems' effects on more languages and conduct more detailed analysis on how the transformer really models the sequence. 

\section{Acknowledgments}
This work is supported by the project "Building a machine translation system to support translation of documents between Vietnamese and Japanese to help managers and businesses in Hanoi approach to Japanese market", No. TC.02-2016-03.

\bibliographystyle{IEEEtran}
\bibliography{iwslt2018-submission}

\begin{thebibliography}{10}
\providecommand{\url}[1]{#1}
\csname url@rmstyle\endcsname
\providecommand{\newblock}{\relax}
\providecommand{\bibinfo}[2]{#2}
\providecommand\BIBentrySTDinterwordspacing{\spaceskip=0pt\relax}
\providecommand\BIBentryALTinterwordstretchfactor{4}
\providecommand\BIBentryALTinterwordspacing{\spaceskip=\fontdimen2\font plus
\BIBentryALTinterwordstretchfactor\fontdimen3\font minus
  \fontdimen4\font\relax}
\providecommand\BIBforeignlanguage[2]{{%
\expandafter\ifx\csname l@#1\endcsname\relax
\typeout{** WARNING: IEEEtran.bst: No hyphenation pattern has been}%
\typeout{** loaded for the language `#1'. Using the pattern for}%
\typeout{** the default language instead.}%
\else
\language=\csname l@#1\endcsname
\fi
#2}}

\bibitem{Sennrich2016a}
R.~Sennrich, B.~Haddow, and A.~Birch, ``{Neural Machine Translation of Rare
  Words with Subword Units},'' in \emph{Association for Computational
  Linguistics (ACL 2016)}, Berlin, Germany, August 2016.

\bibitem{zhangkomachi2018}
\BIBentryALTinterwordspacing
L.~Zhang and M.~Komachi, ``Neural machine translation of logographic language
  using sub-character level information,'' in \emph{Proceedings of the Third
  Conference on Machine Translation: Research Papers}.\hskip 1em plus 0.5em
  minus 0.4em\relax Belgium, Brussels: Association for Computational
  Linguistics, Oct. 2018, pp. 17--25. [Online]. Available:
  \url{https://www.aclweb.org/anthology/W18-6303}
\BIBentrySTDinterwordspacing

\bibitem{Zhang2018}
\BIBentryALTinterwordspacing
J.~Zhang and T.~Matsumoto, ``Improving character-level japanese-chinese neural
  machine translation with radicals as an additional input feature,''
  \emph{CoRR}, vol. abs/1805.02937, 2018. [Online]. Available:
  \url{http://arxiv.org/abs/1805.02937}
\BIBentrySTDinterwordspacing

\bibitem{HochreiterLSTM}
\BIBentryALTinterwordspacing
S.~Hochreiter and J.~Schmidhuber, ``Long short-term memory,'' \emph{Neural
  Comput.}, vol.~9, no.~8, pp. 1735--1780, Nov. 1997. [Online]. Available:
  \url{http://dx.doi.org/10.1162/neco.1997.9.8.1735}
\BIBentrySTDinterwordspacing

\bibitem{Cho2014}
K.~Cho, B.~van Merrienboer, {\c{C}}.~G{\"{u}}l{\c{c}}ehre, F.~Bougares,
  H.~Schwenk, and Y.~Bengio, ``{Learning Phrase Representations using RNN
  Encoder-Decoder for Statistical Machine Translation},'' in \emph{{Proceedings
  of Eighth Workshop on Syntax, Semantics and Structure in Statistical
  Translation (SSST-8}}.\hskip 1em plus 0.5em minus 0.4em\relax Baltimore, ML,
  USA: Association for Computational Linguistics, Jule 2014.

\bibitem{he2016deep}
\BIBentryALTinterwordspacing
K.~He, X.~Zhang, S.~Ren, and J.~Sun, ``Deep residual learning for image
  recognition,'' \emph{CoRR}, vol. abs/1512.03385, 2015. [Online]. Available:
  \url{http://arxiv.org/abs/1512.03385}
\BIBentrySTDinterwordspacing

\bibitem{Luong2015b}
\BIBentryALTinterwordspacing
M.-T. Luong, H.~Pham, and C.~D. Manning, ``Effective approaches to
  attention-based neural machine translation,'' in \emph{Proceedings of the
  2015 Conference on Empirical Methods in Natural Language Processing (EMNLP
  15}.\hskip 1em plus 0.5em minus 0.4em\relax Lisbon, Portugal: Association for
  Computational Linguistics, September 2015, pp. 1412--1421. [Online].
  Available: \url{http://aclweb.org/anthology/D15-1166}
\BIBentrySTDinterwordspacing

\bibitem{VaswaniSPUJGKP17}
\BIBentryALTinterwordspacing
A.~Vaswani, N.~Shazeer, N.~Parmar, J.~Uszkoreit, L.~Jones, A.~N. Gomez,
  L.~Kaiser, and I.~Polosukhin, ``Attention is all you need,'' \emph{CoRR},
  vol. abs/1706.03762, 2017. [Online]. Available:
  \url{http://arxiv.org/abs/1706.03762}
\BIBentrySTDinterwordspacing

\bibitem{VinhNgo2018}
T.-V. Ngo, T.-L. Ha, P.-T. Nguyen, and L.-M. Nguyen, ``Combining advanced
  methods in japanese-vietnamese neural machine translation,'' \emph{2018 10th
  International Conference on Knowledge and Systems Engineering (KSE)}, pp.
  318--322, 2018.

\bibitem{Riza2016}
H.~{Riza}, M.~{Purwoadi}, {Gunarso}, T.~{Uliniansyah}, A.~A. {Ti}, S.~M.
  {Aljunied}, L.~C. {Mai}, V.~T. {Thang}, N.~P. {Thai}, V.~{Chea}, R.~{Sun},
  S.~{Sam}, S.~{Seng}, K.~M. {Soe}, K.~T. {Nwet}, M.~{Utiyama}, and C.~{Ding},
  ``Introduction of the asian language treebank,'' in \emph{2016 Conference of
  The Oriental Chapter of International Committee for Coordination and
  Standardization of Speech Databases and Assessment Techniques (O-COCOSDA)},
  Oct 2016, pp. 1--6.

\bibitem{Neubig2011}
\BIBentryALTinterwordspacing
G.~Neubig, Y.~Nakata, and S.~Mori, ``Pointwise prediction for robust, adaptable
  japanese morphological analysis,'' in \emph{Proceedings of the 49th Annual
  Meeting of the Association for Computational Linguistics: Human Language
  Technologies: Short Papers - Volume 2}, ser. HLT '11.\hskip 1em plus 0.5em
  minus 0.4em\relax Stroudsburg, PA, USA: Association for Computational
  Linguistics, 2011, pp. 529--533. [Online]. Available:
  \url{http://dl.acm.org/citation.cfm?id=2002736.2002841}
\BIBentrySTDinterwordspacing

\bibitem{opennmt}
G.~Klein, Y.~Kim, Y.~Deng, J.~Senellart, and A.~M. Rush, ``Opennmt: Open-source
  toolkit for neural machine translation,'' in \emph{Proceedings of the 55th
  Annual Meeting of the Association for Computational Linguistics-System
  Demonstrations}.\hskip 1em plus 0.5em minus 0.4em\relax Vancouver, Canada,
  July 30 - August 4, 2017: Association for Computational Linguistics, 2017,
  pp. 67--72.

\bibitem{Isozaki2010}
\BIBentryALTinterwordspacing
H.~Isozaki, T.~Hirao, K.~Duh, K.~Sudoh, and H.~Tsukada, ``Automatic evaluation
  of translation quality for distant language pairs,'' in \emph{Proceedings of
  the 2010 Conference on Empirical Methods in Natural Language Processing},
  ser. EMNLP '10.\hskip 1em plus 0.5em minus 0.4em\relax Stroudsburg, PA, USA:
  Association for Computational Linguistics, 2010, pp. 944--952. [Online].
  Available: \url{http://dl.acm.org/citation.cfm?id=1870658.1870750}
\BIBentrySTDinterwordspacing

\bibitem{Lee2016}
\BIBentryALTinterwordspacing
J.~Lee, K.~Cho, and T.~Hofmann, ``Fully character-level neural machine
  translation without explicit segmentation,'' \emph{CoRR}, vol.
  abs/1610.03017, 2016. [Online]. Available:
  \url{http://arxiv.org/abs/1610.03017}
\BIBentrySTDinterwordspacing

\bibitem{Cherry2018}
\BIBentryALTinterwordspacing
C.~Cherry, G.~Foster, A.~Bapna, O.~Firat, and W.~Macherey, ``Revisiting
  character-based neural machine translation with capacity and compression,''
  \emph{CoRR}, vol. abs/1808.09943, 2018. [Online]. Available:
  \url{http://arxiv.org/abs/1808.09943}
\BIBentrySTDinterwordspacing

\end{thebibliography}

\end{document}